\documentclass[runningheads]{llncs}

\usepackage{amsmath}
\usepackage[T1]{fontenc}
\usepackage{amssymb}
\usepackage{float}
\usepackage{wrapfig}
\usepackage{booktabs}
\usepackage{graphicx}
\usepackage{adjustbox}
\usepackage{multirow}
\usepackage{xcolor}
\usepackage{soul}
\usepackage{hyperref}
\begin{document}
%
\title{SNAP-KG: Streaming Node Assignment via Projection for Knowledge Graph Entity Integration}
%
\titlerunning{Streaming Node Assignment via Projection for KG Entity Integration}
%
\author{Jui-Chien Lin\inst{1}\orcidID{0009-0000-7640-1140} \and
Mohammad Mohammadi Amiri\inst{1}\orcidID{0000-0001-6164-9020} \and
Oshani Seneviratne\inst{1}\orcidID{0000-0001-8518-917X}}
\authorrunning{J.-C. Lin, Amiri, and Seneviratne}
\institute{Rensselaer Polytechnic Institute, Troy, NY 12180, USA\\
\email{\{linj26, mamiri, senevo\}@rpi.edu}}
\maketitle              
\begin{abstract}
Knowledge graph (KG) construction pipelines must continuously integrate
newly arriving entities into a growing graph. Unlike the commonly studied
setting of inserting triples between already-present nodes, a newly arriving
entity carries no graph connectivity whatsoever: it emerges from the upstream
acquisition phase as a raw feature vector and must be assigned to a semantic
community before entity resolution and link prediction can be executed against
a tractably scoped candidate set rather than the entire node population.
Existing multi-view graph clustering methods produce high-quality, relationally
informed community structure by exploiting multiple distinct relation types as
separate structural views, yet every such method is transductive: it assumes a
complete, fixed graph during training and cannot assign unseen entities to
clusters without retraining from scratch, a cost incompatible with continuous
streaming ingestion. We propose \textbf{SNAP-KG} (Streaming Node Assignment
via Projection for Knowledge Graph Entity Integration), a framework that
simultaneously supports \emph{graph-structural} multi-view relational
clustering, where views are defined by distinct KG relation types rather
than feature modalities, and inductive inference for streaming entities.
SNAP-KG trains a projector network to map any newly arriving entity directly
to the learned embedding space using only its raw features, enabling immediate
cluster assignment without graph access or model retraining, reducing inference
from a training-time operation to a single feed-forward pass. Experiments on
five benchmark multi-view graph datasets and a production-scale KG of 2.4
million nodes demonstrate multiple orders-of-magnitude inference speedups over
retraining-based approaches and competitive clustering quality despite the
inductive design. SNAP-KG is further evaluated as a candidate scoping
mechanism for downstream tasks: the induced clusters achieve 62--75\% candidate search reduction on the five benchmark datasets and
97\% on OGB-WikiKG2 for entity resolution and link prediction.


\end{abstract}

\section{Introduction}

\begin{figure}[t]
\centering
\includegraphics[width=0.8\textwidth]{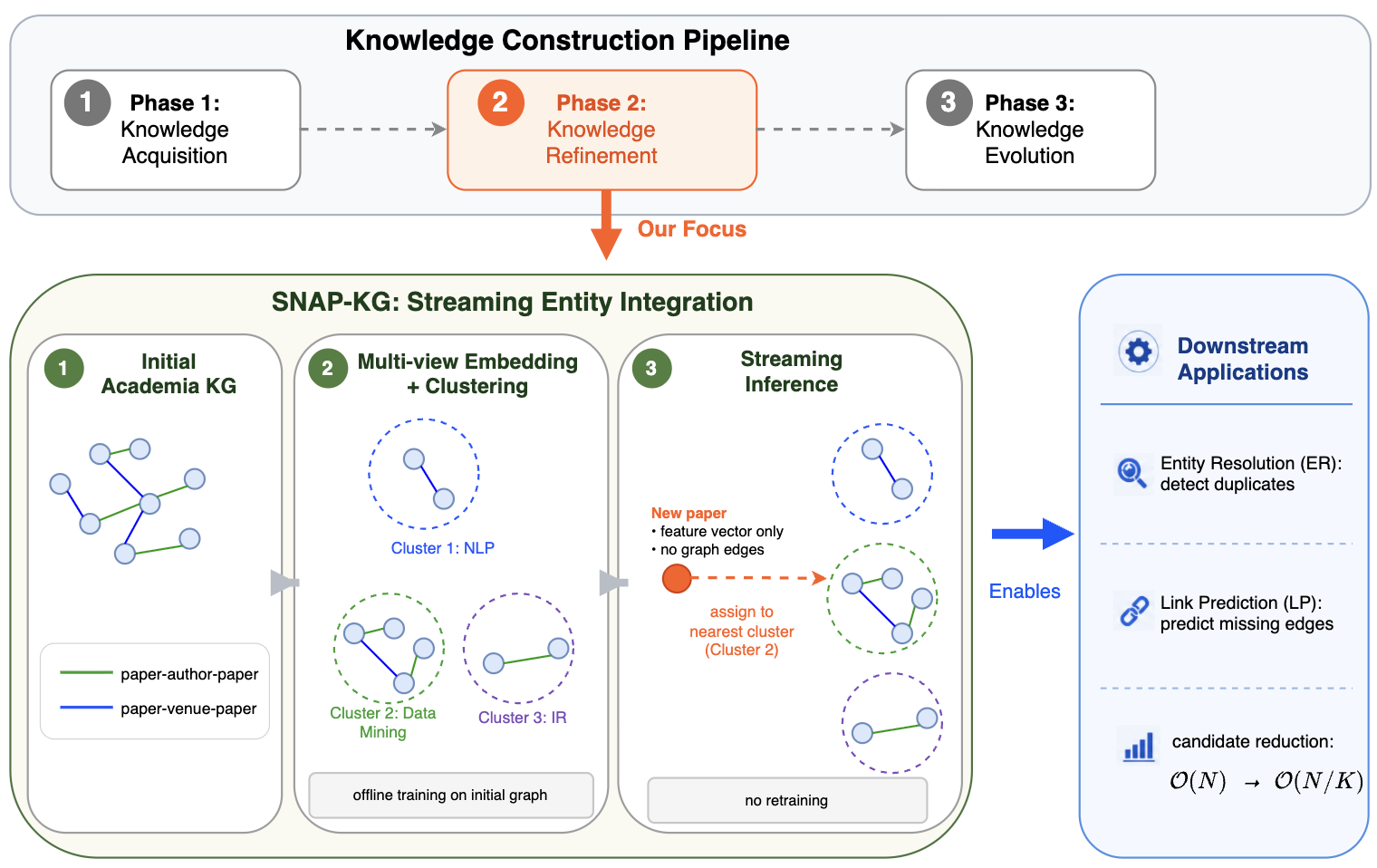}
\caption{SNAP-KG addresses the streaming entity integration sub-problem within
Phase~2 (Knowledge Refinement) of the KG construction pipeline. A newly
arriving entity carries only raw node features and no graph connectivity.
SNAP-KG assigns it to an existing cluster via an inductive projector
$f_{\phi}$ without retraining, enabling downstream entity resolution~(ER) and
link prediction~(LP) to be scoped within the assigned cluster and reducing the
candidate search space.}
\label{fig:position}
\end{figure}

Knowledge graphs (KGs) encode world knowledge as semantically typed triples,
underpinning applications from question answering to information
retrieval~\cite{hogan2021knowledge}. Their construction follows a three-phase
pipeline~\cite{zhong2023comprehensive}: \textbf{Knowledge Acquisition} converts
raw text into an initial graph via named entity recognition and relation
extraction; \textbf{Knowledge Refinement} merges duplicates via entity
resolution (ER) and completes missing facts via link prediction
(LP)~\cite{paulheim2016knowledge}; and \textbf{Knowledge Evolution} tracks
fact changes over time~\cite{weikum2021machine}. Extending this pipeline to
streaming settings introduces a structural growth problem orthogonal to
Knowledge Evolution: integrating entities new to the graph, as shown in
Figure~\ref{fig:position}.

Each newly arriving entity emerges from the acquisition phase as a \emph{raw
feature vector with no graph connectivity}: it has no edges, no neighbors, and
no structural context. Before ER or LP can be applied, this entity must be
placed into a semantically coherent scope so that downstream tasks operate over
a tractable candidate set. Naive application of ER and LP incurs an
$\mathcal{O}(N)$ bottleneck per entity: ER must scan all $N$ existing
nodes~\cite{getoor2012entity} and LP must score all $N$ candidate
targets, prohibitive at real-world KG scale~\cite{hogan2021knowledge}. Graph
clustering addresses this by partitioning the node set into $K$ communities,
reducing the per-entity cost to $\mathcal{O}(N/K)$. Unlike conventional
blocking~\cite{papadakis2020blocking}, which partitions records via
attribute-space keys in isolation from graph
structure~\cite{thirumuruganathan2021deep}, graph-based partitions encode
relational similarity so that candidate pairs share meaningful topological
context~\cite{fortunato2010community}, and cluster membership independently
supports entity recommendation and exploratory graph
analytics~\cite{fortunato2010community,shao2019community,su2022comprehensive}.

\paragraph{Problem Statement.}
In a KG, each distinct relation type constitutes one \emph{view}, represented by its own adjacency matrix over the shared node set. We are given a multi-view
KG snapshot $G_0$ comprising a node set $\mathcal{V}$ with $|\mathcal{V}|{=}N$,
$M$ such views each with adjacency matrix
$\boldsymbol{A}^{(m)}\!\in\!\mathbb{R}^{N\times N}$, and a shared node feature
matrix $\boldsymbol{X}\!\in\!\mathbb{R}^{N\times D}$. Entities arrive as a
stream, each as a raw feature vector
$\boldsymbol{x}_{\mathrm{new}}\!\in\!\mathbb{R}^{D}$ with \emph{no edges in
any view}. The goal is to assign each $\boldsymbol{x}_{\mathrm{new}}$ to one
of $K$ semantically coherent clusters \emph{without model retraining and
without graph access at inference time}, such that the assigned cluster scopes
subsequent ER and LP. Three factors make this jointly difficult:
\textbf{(i)}~a cold-start barrier, as no neighborhood exists for Graph Neural Network (GNN)
aggregation; \textbf{(ii)}~a latency constraint, as per-batch retraining is
orders of magnitude too slow for high-throughput streaming; and
\textbf{(iii)}~a multi-view requirement, as no single relation type suffices
for discriminative clustering.

\paragraph{The Gap.}
This graph-structural notion of view differs from the modality-based definition
in the broader multi-view learning literature~\cite{yang2018multi}, where views
denote heterogeneous feature sources such as text or images. Different relation
types carry complementary structural signals, and fusing them yields more
discriminative representations than any single view
alone~\cite{pan2021multi,hassani2020contrastive}. Yet multi-view graph
clustering
methods~\cite{pan2021multi,shen2024balanced,mo2023disentangled,ling2023dual,lin2025effective}
are \emph{transductive}: they require the complete graph during training and
cannot embed unseen nodes without full retraining, violating
requirements~(i) and~(ii). Inductive GNN
frameworks~\cite{hamilton2017inductive,velivckovic2017graph,hu2020heterogeneous}
handle unseen nodes but operate on single-view homogeneous graphs,
ignoring~(iii). Neural blocking methods~\cite{thirumuruganathan2021deep}
support inductive candidate generation but discard relational structure
entirely, likewise failing~(iii). Beyond these camps, inductive methods for
modality-based multi-view clustering~\cite{wei2021inductive} and streaming
methods for incrementally arriving views~\cite{yuan2025deep} do exist, but
both operate exclusively over per-node feature spaces and never leverage
graph-structural relation types at any stage, neither training nor
inference involves adjacency matrices or relational topology. Consequently,
neither addresses the core challenge of this work: learning community
structure that is \emph{informed by multi-view relational topology} and
then deploying it inductively for streaming entities arriving without graph
connectivity. To the best of our knowledge, no existing method simultaneously
satisfies all three requirements in the graph-structural, streaming-entity
setting.

We propose \textbf{SNAP-KG} (\textbf{S}treaming \textbf{N}ode \textbf{A}ssignment
via \textbf{P}rojection for \textbf{K}nowledge \textbf{G}raph Entity
Integration), a streaming refinement plug-in that closes this gap. SNAP-KG is
not an end-to-end KG construction system; it takes entity representations from
the upstream acquisition phase and outputs community assignments scoping
subsequent ER and LP. It is therefore agnostic to how the KG was built. In the \textbf{offline training} phase (run once on
$G_0$), view-specific GNNs produce per-relation embeddings that a Transformer
encoder fuses into unified, cluster-discriminative representations via
contrastive learning. A lightweight Multi-Layer Perceptron (MLP) projector $f_\phi$ is simultaneously
trained to reconstruct these fused embeddings from raw node features alone,
following the GNN-to-MLP knowledge distillation
paradigm~\cite{zhang2021graph}. During \textbf{online inference}, any
$\boldsymbol{x}_{\mathrm{new}}$ is mapped by $f_\phi$ and assigned to the nearest cluster
centroid with no graph access and no retraining.

The primary benefit is eliminating model retraining entirely: retraining
dominates per-batch latency, whereas cluster assignment is orders of magnitude
faster. We use $K$-means in this work, though the framework is agnostic to the
choice of clustering algorithm. Because re-clustering is far cheaper than
retraining, the cluster structure can optionally be refreshed to adapt to
evolving graph distributions without modifying model weights. The $K$ graph-aware cluster centroids are further composable with
Approximate Nearest Neighbor (ANN) indices based on Inverted File
Index (IVF) structures~\cite{johnson2019billion} for production-scale
retrieval; we leave the evaluation of such composed pipelines to
future work.


The main contributions of this work are as follows:
\begin{itemize}

\item \textbf{First inductive \emph{graph-structural} multi-view framework
for streaming KG entity integration.} Existing approaches fall into two
disjoint camps: multi-view graph clustering methods requiring full retraining
per batch, and inductive GNN methods operating on single-view graphs.
SNAP-KG bridges this gap via an offline multi-view training phase whose
graph-aware embeddings are deployed inductively at zero retraining cost.

\item \textbf{GNN-to-MLP distillation for inductive multi-view graph clustering.}
 While the
paradigm~\cite{zhang2021graph} is established in single-view settings, no prior
work applies it to a multi-view relational teacher, a fused GNN-Transformer,
targeting community-aware entity assignment.

\item \textbf{Empirical validation across clustering and downstream scoping.}
SNAP-KG's inductive projector matches retrain-level cluster assignment
quality at multiple orders-of-magnitude lower inference cost, while the full
model achieves competitive clustering quality against transductive
state-of-the-art baselines across five multi-view graph benchmarks
(ACM~\cite{ren2025multi}, DBLP~\cite{ren2025multi}, IMDB~\cite{ren2025multi}; YELP~\cite{lin2025effective}, MAG~\cite{lin2025effective})
and a 2.4M-node production KG~\cite{hu2020open}, with
62\,\%--75\,\% candidate search reduction for downstream ER and
relation-agnostic edge prediction scoping.
\end{itemize}

\section{Related Work}

\subsection{Multi-View Graph Clustering}

Graph clustering exploits topological connectivity and node attributes to form
semantically coherent communities~\cite{fortunato2010community}. In multi-view
settings where each distinct relation type induces a separate adjacency matrix,
methods range from subspace-based and consensus-graph
approaches~\cite{cao2015diversity,kang2020large,nie2017self} to deep contrastive
frameworks such as O2MAC~\cite{fan2020one2multi}, MCGC~\cite{pan2021multi},
DuaLGR~\cite{ling2023dual}, MGDCR~\cite{mo2023multiplex},
DMG~\cite{mo2023disentangled}, BMGC~\cite{shen2024balanced}, and the scalable
DEMM~\cite{lin2025effective}. Despite strong clustering quality, all are
\emph{transductive}: training requires the complete node set and all adjacency
matrices, and learned embeddings cannot be extended to unseen nodes without full
retraining~\cite{pan2021multi,hamilton2017inductive}, incompatible with
streaming ingestion. SNAP-KG inherits multi-view relational quality while
eliminating this dependency through knowledge distillation.

\subsection{Inductive Graph Representation Learning}

Inductive methods embed previously unseen nodes without retraining.
GraphSAGE~\cite{hamilton2017inductive} learns neighborhood aggregation
functions; HGT~\cite{hu2020heterogeneous} extends message passing with
query-key attention for multi-relational graphs. The GNN-to-MLP distillation
paradigm~\cite{zhang2021graph,zhou2024tined} eliminates graph access at
inference by training an MLP student over frozen GNN teacher embeddings.
However, existing GNN-to-MLP distillation methods have only been evaluated with single-view GNN architectures. Inductive
methods for modality-based multi-view clustering~\cite{yang2025towards,wei2021inductive}
address cross-modal generalization over heterogeneous feature sources rather
than the graph-structural relation types studied here. Streaming multi-view
clustering methods~\cite{yuan2025deep} handle incrementally arriving feature
modalities over a fixed sample set, not streaming entities arriving without
graph connectivity. SNAP-KG is the first to apply GNN-to-MLP distillation to a
multi-view relational teacher targeting community-aware entity assignment,
satisfying all three requirements simultaneously.

\subsection{Streaming Knowledge Graph Construction}
Extending KG construction to continuous data streams raises challenges related to
incremental entity integration, memory efficiency, and operational continuity for
downstream querying~\cite{weikum2021machine}. A major family of streaming KGC tools
operates at the data-lifting layer, converting structured or semi-structured source
streams into RDF triples via declarative mapping rules.
RMLStreamer~\cite{oo2022rmlstreamer} is a scalable parallel engine that executes RML
rules over heterogeneous data streams, delivering high throughput at low latency with
constant memory consumption. RMLWeaver-JS~\cite{oo2024rmlweaver} is a complementary reactive algebraic engine that
translates RML documents into composable operator plans, likewise demonstrating constant
memory usage across varying workloads. Together, these systems show that declarative
stream-to-RDF lifting can operate at high throughput and low latency. A second
family instead builds KGs from unstructured text with large language models, as in
CoDe-KG~\cite{anuyah2025automated}.
SNAP-KG occupies a complementary and architecturally distinct position with respect to
both: rather than
generating RDF triples from mapping rules or extracting them from text, it operates downstream of these layers
on entity feature representations already extracted from the data stream, assigning
newly arriving entities to existing semantic communities so that ER and LP can be
executed with reduced candidate sets. RMLStreamer and RMLWeaver handle the
stream-to-RDF layer and LLM-based systems the extraction from text; SNAP-KG
handles community-aware entity integration within the
already-constructed graph, forming composable stages of the same broader
pipeline~\cite{zhong2023comprehensive}.

\subsection{Blocking for Entity Resolution and Link Prediction}

ER relies on blocking to prune the $\mathcal{O}(N^2)$ candidate space while
preserving recall over true duplicate pairs~\cite{papadakis2020blocking}; an
analogous scoping problem arises in LP, where scoring all $\mathcal{O}(N)$
target nodes per query is intractable. Neural approaches such as
DeepBlocker~\cite{thirumuruganathan2021deep} embed records via pre-trained
language models for approximate nearest-neighbor retrieval, achieving strong
recall on attribute-rich datasets. However, both traditional and neural blocking
operate solely on node attributes, remaining blind to relational graph structure
and producing partitions with no guaranteed semantic
coherence~\cite{fortunato2010community}. SNAP-KG's cluster-based scoping
encodes multi-view relational similarity, producing semantically coherent
partitions that attribute-only methods fundamentally cannot achieve.

\section{Methodology}
\label{sec:method}

This section details the SNAP-KG framework, covering how view-specific information is
aggregated and fused into unified embeddings, how newly arriving entities are inductively
embedded without graph structure access, and how the framework integrates into streaming
KG construction pipelines.

\subsection{Notation and Node Features}
\label{sec:features}
We define a multi-view graph consisting of $M$ distinct views. Each view $m$ has its own
adjacency matrix $\boldsymbol{A}^{(m)} \in \mathbb{R}^{N \times N}$, where $N$ denotes
the total number of nodes (with $|\mathcal{V}| = N$ and $\mathcal{V}$ the node set). All
views share a common node feature matrix $\boldsymbol{X} \in \mathbb{R}^{N \times D}$,
where $D$ denotes the input feature dimension.

Each row $\boldsymbol{x}_i$ of $\boldsymbol{X}$ is built only from information about
entity $i$, including its surface form, textual description, literal values, and type
labels. No graph-derived statistic such as node degree or predicate count is involved, so relational structure enters training only through the $\boldsymbol{A}^{(m)}$. This design fits the streaming setting, where a newly arriving
entity does not yet have graph statistics. It only requires the same feature extractor
to generate $\boldsymbol{X}$ during training and $\boldsymbol{x}_{\mathrm{new}}$ during
inference. ACM, DBLP, IMDB, YELP, and MAG use the feature matrices provided with the
original datasets, while OGB-WikiKG2 uses sentence-transformer
embeddings~\cite{reimers2019sentence} of entity labels and descriptions.
Extraction-based and LLM-based pipelines naturally produce these features. For a
traditional knowledge graph that contains only URIs, an additional feature extraction
step from entity labels and literal values is needed.

\subsection{Adaptive Graph Preprocessing}

Real-world KG adjacency matrices often contain noisy or weakly informative
edges induced by hub nodes or co-occurrence-based relation construction,
where a node's neighborhood may include semantically unrelated
entities~\cite{ren2025multi}. Aggregating over such neighbors can degrade
view-specific representations, subsequently harming both clustering quality
and projector learning.

To mitigate this, SNAP-KG performs an adaptive neighbor filtering step on
each view prior to training. Let $\mathcal{N}_i^{(m)}$ denote the original
neighborhood of node $i$ under view $m$ defined by $\boldsymbol{A}^{(m)}$,
let $\boldsymbol{X} \in \mathbb{R}^{N \times D}$ be the shared node feature
matrix, and let $\boldsymbol{x}_i \in \mathbb{R}^{D}$ be the feature vector
of node $i$. For each neighbor $j \in \mathcal{N}_i^{(m)}$, we compute the
cosine similarity:

\begin{equation}
s_{ij}^{(m)}
=
\frac{\boldsymbol{x}_i^\top \boldsymbol{x}_j}
{\|\boldsymbol{x}_i\|_2 \|\boldsymbol{x}_j\|_2}.
\end{equation}

Here $s_{ij}^{(m)}$ measures the feature-space proximity between node $i$
and neighbor $j$, and serves as a proxy for edge informativeness: semantically
relevant neighbors tend to exhibit higher feature similarity, whereas noisy
or spurious connections are typically dissimilar in feature space.

We examine the distribution of $\{s_{ij}^{(m)}\}_{j \in \mathcal{N}_i^{(m)}}$
within node $i$'s neighborhood. If most neighbors are similarly relevant,
the distribution is approximately uniform; if only a small subset is strongly
aligned with node $i$ while the rest are weakly related, the distribution
becomes skewed with a long tail of low-similarity scores. The standard
deviation $\sigma_i^{(m)} = \mathrm{std}(\{s_{ij}^{(m)}\}_{j \in
\mathcal{N}_i^{(m)}})$ quantifies this behavior and determines the
filtered neighborhood:

\begin{equation}
\widetilde{\mathcal{N}}_i^{(m)}
=
\begin{cases}
\operatorname{top-}k_u
\left( \mathcal{N}_i^{(m)} \right),
& \text{if } \sigma_i^{(m)} \leq \sigma_{\mathrm{thresh}}, \\[6pt]
\operatorname{top-}
\left\lfloor \rho \cdot |\mathcal{N}_i^{(m)}| \right\rfloor
\left( \mathcal{N}_i^{(m)} \right),
& \text{if } \sigma_i^{(m)} > \sigma_{\mathrm{thresh}},
\end{cases}
\end{equation}

where neighbors are ranked by $s_{ij}^{(m)}$ in descending order.
Here $k_u$ is the fixed neighbor budget for the uniform case, $\rho \in
(0,1]$ is the retention ratio for the skewed case, and $\sigma_{\mathrm{thresh}}$
controls the transition between the two regimes. When the distribution is
approximately uniform, a fixed-size selection avoids retaining unnecessarily
large neighborhoods; when skewed, the ratio-based strategy preserves the
most informative neighbors while discarding weakly related connections.
The filtered adjacency matrices $\widetilde{\boldsymbol{A}}^{(m)}$ are used
throughout all subsequent training stages, and only there: no triple is deleted
and no predicate is dropped, so the stored KG and its semantics are unchanged.
Filtering likewise plays no role at inference.

\subsection{View-Specific Embedding Generation}

For each view $m$, a view-specific GNN, $\mathrm{GNN}^{(m)}$, aggregates information from
the filtered neighborhood $\widetilde{\mathcal{N}}_i^{(m)}$ to produce view-specific node
embeddings:

\begin{equation}
\boldsymbol{h}_i^{(m)} = \mathrm{GNN}^{(m)}\!\left(
\boldsymbol{x}_i,\;
\left\{ \boldsymbol{x}_j \mid j \in \widetilde{\mathcal{N}}_i^{(m)} \right\}
\right).
\end{equation}

\begin{figure}[t]
\centering
\includegraphics[width=0.8\textwidth]{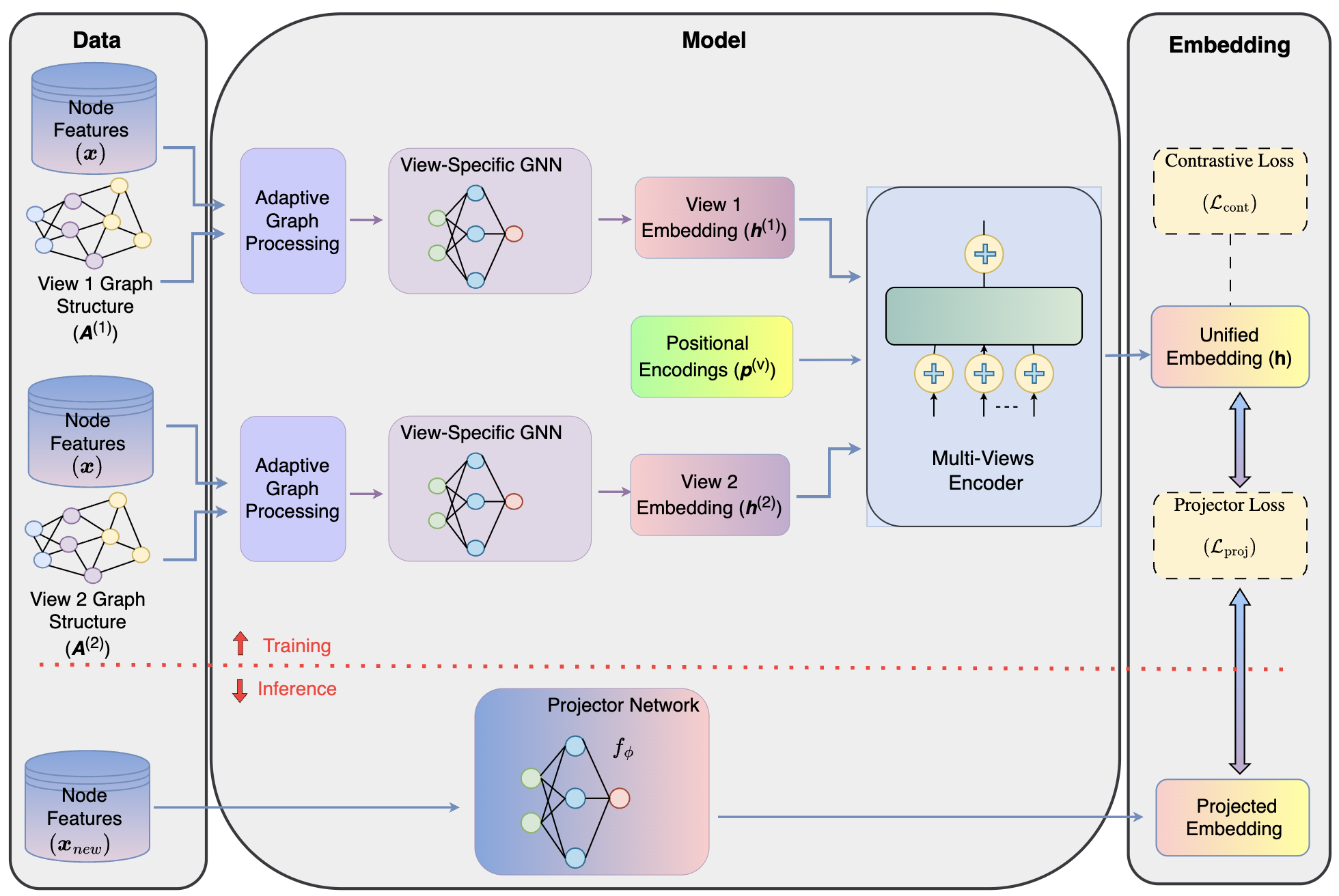}
\caption{Overview of SNAP-KG framework. The projector network is trained to simulate the
combined functionality of the view-specific GNNs and multi-view encoder, enabling
efficient embedding generation for streaming nodes. For clarity, only two views are
shown.}
\label{fig:model_demo}
\end{figure}

Each edge between $\boldsymbol{x}_i$ and $\boldsymbol{x}_j$ encodes a view-specific
correlation, enriching the representation of node $i$ with complementary contextual
information from its local neighborhood, as illustrated in Figure~\ref{fig:model_demo}.

\subsection{Multi-View Aggregation via Encoder}

The view-specific embeddings are aggregated into a unified representation using a
transformer-based encoder~\cite{devlin2019bert}. Each view embedding is augmented with a
learnable positional encoding $\boldsymbol{p}^{(m)}$ to distinguish between views, and a
special aggregation token $\boldsymbol{z}_{\mathrm{agg}}$ is prepended to the sequence:

\begin{equation}
\boldsymbol{h}_i
=
\mathrm{Encoder}\big[
\boldsymbol{z}_{\mathrm{agg}},\,
\boldsymbol{h}^{(1)}_i + \boldsymbol{p}^{(1)},\,
\boldsymbol{h}^{(2)}_i + \boldsymbol{p}^{(2)},\,
\dots,\,
\boldsymbol{h}^{(M)}_i + \boldsymbol{p}^{(M)}
\big]_{[0]},
\end{equation}

\noindent where $[\cdot]_{[0]}$ denotes the output at the aggregation token position.
The self-attention mechanism allows $\boldsymbol{z}_{\mathrm{agg}}$ to attend to all
views, yielding the unified embedding $\boldsymbol{h}_i$.

\subsection{Contrastive Learning}

We adopt contrastive learning to encourage cluster-separable embeddings.
The positive neighborhood of node $i$ is the union of its filtered neighborhoods
across all views:

\begin{equation}
\begin{aligned}
\mathcal{N}_i^{+}
&=
\bigcup_{m=1}^{M} \widetilde{\mathcal{N}}_i^{(m)}, \\
\mathcal{N}_i^{-}
&=
\mathcal{V} \setminus \left( \{i\} \cup \mathcal{N}_i^{+} \right).
\end{aligned}
\end{equation}

The node-level contrastive loss uses a temperature-scaled similarity metric:

\begin{equation}
\mathcal{L}_i = -\log \frac{\sum_{j \in \mathcal{N}^{+}_i} \exp(\mathrm{sim}(\boldsymbol{h}_i, \boldsymbol{h}_j)/\tau)}{\sum_{k \neq i} \exp(\mathrm{sim}(\boldsymbol{h}_i, \boldsymbol{h}_k)/\tau)},
\end{equation}

\noindent where $\mathrm{sim}(\cdot,\cdot)$ denotes cosine similarity and $\tau$ is the
temperature parameter. The overall contrastive loss is:

\begin{equation}
\mathcal{L}_{\mathrm{cont}} = \frac{1}{N} \sum_{i \in \mathcal{V}} \mathcal{L}_i.
\end{equation}

\subsection{Projector Network}

To enable inductive embedding for streaming entities that arrive without neighborhood
information, we introduce a projector network following the GNN-to-MLP knowledge
distillation paradigm~\cite{zhang2021graph}. The projector $f_{\phi}$, a MLP, minimizes the squared L2 reconstruction loss between projected and fused
embeddings:
\begin{equation}
\mathcal{L}_{\mathrm{proj}} = \frac{1}{N} \sum_{i=1}^{N} \|\boldsymbol{h}_i - f_{\phi}(\boldsymbol{x}_i)\|^2_2.
\end{equation}

This allows high-quality embeddings to be generated for streaming nodes using only raw
node features, without requiring graph structure or retraining.

\subsection{Joint Optimization}

The full training objective combines the contrastive and projector losses:

\begin{equation}
\mathcal{L}_{\mathrm{total}} = \mathcal{L}_{\mathrm{cont}} + \lambda \mathcal{L}_{\mathrm{proj}},
\end{equation}

\noindent where $\lambda$ balances the two objectives. This encourages that: (i) the
view-specific GNN encoders capture neighborhood structure within each relation type;
(ii) the Transformer encoder integrates multi-view embeddings via cross-view attention;
and (iii) the projector accurately approximates the full GNN-Transformer pipeline from
raw node features alone.

\subsection{Streaming Inference and KG Construction Integration}

SNAP-KG is trained once on the initial multi-view KG, where each view corresponds to a
distinct relation type. During inference, a new entity with feature vector
$\boldsymbol{x}_{\mathrm{new}}$ is embedded via the projector,
$\boldsymbol{h}_{\mathrm{new}} = f_{\phi}(\boldsymbol{x}_{\mathrm{new}})$, and assigned
to the nearest cluster centroid without retraining or graph reconstruction.

The resulting cluster assignments serve as a scope-reduction mechanism for two core KG
construction tasks. In \textbf{entity resolution}, duplicate detection is restricted to
intra-cluster candidates, reducing per-entity comparison cost from $\mathcal{O}(N)$ to
$\mathcal{O}(N/K)$ where $K$ is the number of clusters~\cite{papadakis2020blocking}. In
\textbf{link prediction}, confining candidate edge inference to intra-community pairs
provides a computationally tractable and semantically coherent scope for the expanding
graph. Section~\ref{sec:ER_LP} evaluates both tasks empirically, showing that
the induced cluster structure substantially reduces candidate search cost while
keeping task quality close to the full-scan upper bound.

\section{Experiment}
We evaluate SNAP-KG across four dimensions. \textbf{(1) Streaming inference efficiency}
(Section~\ref{sec:inductive_exp}): does the projector match retrain-level
cluster assignment quality, and at what inference cost relative to
retraining baselines? \textbf{(2) Clustering quality}
(Section~\ref{sec:MVGC}): how does SNAP-KG's inductive design compare
against transductive state-of-the-art methods on standard benchmarks?
\textbf{(3) Ablation} (Section~\ref{sec:ablation}): is multi-view fusion
necessary? \textbf{(4) Downstream scoping utility}
(Sections~\ref{sec:ER_LP}--\ref{sec:wikikg2}): does the induced cluster
structure provide meaningful candidate reduction for ER and LP without
degrading task quality?


\begin{table}[!htb]
    \centering
    \small
    \caption{Statistics of the benchmark datasets.}
    \label{tab:dataset_intro}
    \begin{adjustbox}{max width=0.6\textwidth}
\begin{tabular}{lrlrr}
\hline
Dataset &
  \multicolumn{1}{l}{Nodes} &
  Views (Meta-paths) &
  \multicolumn{1}{l}{Features} &
  \multicolumn{1}{l}{Clusters} \\ \hline
\multicolumn{1}{l|}{\multirow{2}{*}{ACM}} &
  \multicolumn{1}{r|}{\multirow{2}{*}{3025}} &
  \multicolumn{1}{l|}{Paper-Author-Paper} &
  \multicolumn{1}{r|}{\multirow{2}{*}{1870}} &
  \multirow{2}{*}{3} \\
\multicolumn{1}{l|}{} &
  \multicolumn{1}{r|}{} &
  \multicolumn{1}{l|}{Paper-Subject-Paper} &
  \multicolumn{1}{r|}{} &
   \\ \hline
\multicolumn{1}{l|}{\multirow{3}{*}{DBLP}} &
  \multicolumn{1}{r|}{\multirow{3}{*}{4057}} &
  \multicolumn{1}{l|}{Author-Paper-Author} &
  \multicolumn{1}{r|}{\multirow{3}{*}{334}} &
  \multirow{3}{*}{4} \\
\multicolumn{1}{l|}{} &
  \multicolumn{1}{r|}{} &
  \multicolumn{1}{l|}{Author-Paper-Venue-Paper-Author} &
  \multicolumn{1}{r|}{} &
   \\
\multicolumn{1}{l|}{} &
  \multicolumn{1}{r|}{} &
  \multicolumn{1}{l|}{Author-Paper-Term-Paper-Author} &
  \multicolumn{1}{r|}{} &
   \\ \hline
\multicolumn{1}{l|}{\multirow{2}{*}{IMDB}} &
  \multicolumn{1}{r|}{\multirow{2}{*}{4780}} &
  \multicolumn{1}{l|}{Movie-Actor-Movie} &
  \multicolumn{1}{r|}{\multirow{2}{*}{1232}} &
  \multirow{2}{*}{3} \\
\multicolumn{1}{l|}{} &
  \multicolumn{1}{r|}{} &
  \multicolumn{1}{l|}{Movie-Director-Movie} &
  \multicolumn{1}{r|}{} &
   \\ \hline
\multicolumn{1}{l|}{\multirow{3}{*}{YELP}} &
  \multicolumn{1}{r|}{\multirow{3}{*}{2614}} &
  \multicolumn{1}{l|}{Business-Location-Business} &
  \multicolumn{1}{r|}{\multirow{3}{*}{82}} &
  \multirow{3}{*}{3} \\
\multicolumn{1}{l|}{} &
  \multicolumn{1}{r|}{} &
  \multicolumn{1}{l|}{Business-Service-Business} &
  \multicolumn{1}{r|}{} &
   \\
\multicolumn{1}{l|}{} &
  \multicolumn{1}{r|}{} &
  \multicolumn{1}{l|}{Business-User-Business} &
  \multicolumn{1}{r|}{} &
   \\ \hline
\multicolumn{1}{l|}{\multirow{2}{*}{MAG}} &
  \multicolumn{1}{r|}{\multirow{2}{*}{113919}} &
  \multicolumn{1}{l|}{Paper-Paper} &
  \multicolumn{1}{r|}{\multirow{2}{*}{128}} &
  \multirow{2}{*}{4} \\
\multicolumn{1}{l|}{} &
  \multicolumn{1}{r|}{} &
  \multicolumn{1}{l|}{Paper-Author-Paper} &
  \multicolumn{1}{r|}{} &
   \\ \hline
\end{tabular}%

    \end{adjustbox}
\end{table}

We evaluate SNAP-KG on five benchmark multi-view graph datasets: \textbf{ACM},
\textbf{DBLP}, \textbf{IMDB}, \textbf{YELP}, and \textbf{MAG}. Table~\ref{tab:dataset_intro}
summarizes the statistics of these datasets, including node count, feature dimensionality,
number of views, and number of classes. Class labels serve as ground truth for
clustering evaluation.

\paragraph{Choosing $K$.}
On the five benchmarks, we set $K$ to the number of ground-truth classes, following the
standard protocol~\cite{pan2021multi,shen2024balanced,lin2025effective} for a fair
comparison in Section~\ref{sec:MVGC}. In deployment, ground-truth classes are unavailable,
so $K$ must be chosen differently. It controls a trade-off: candidate size decreases as
$\mathcal{O}(N/K)$, while the chance that a true neighbor falls into a different cluster
increases with $K$. In practice, $K$ can be selected from a latency budget, tuned on a
validation set, or, without labels, by an internal criterion such as the
silhouette score~\cite{rousseeuw1987silhouettes}. In
Section~\ref{sec:wikikg2}, we sweep $K \in [20,2000]$ and use the stable range
$\{200,225,250,275\}$, where ER F1 varies by only 1.3 points.

\subsection{Streaming Inference: Speed and Quality}
\label{sec:inductive_exp}
The central efficiency claim of SNAP-KG is that the projector $f_{\phi}$
eliminates per-batch retraining entirely, reducing inference to a single
feed-forward pass. This experiment verifies that this speedup comes at no
quality cost by comparing four inference strategies. \textbf{Isolated} forwards
$\boldsymbol{x}_{\mathrm{new}}$ through the GNN-Transformer pipeline without
neighborhood context; since no neighbors are available, GNN aggregation
degenerates to a per-node MLP pass over raw features alone, discarding all
relational structure. \textbf{Neighbors} retrieves pseudo-neighbors via KNN
before applying the full GNN pipeline; since KNN operates on feature similarity
without regard to relation type, the neighborhood context is relation-agnostic
and multi-view structure is not exploited. \textbf{SNAP-KG-Projector} maps
$\boldsymbol{x}_{\mathrm{new}}$ directly to the embedding space via $f_{\phi}$
alone, requiring no graph access. \textbf{SNAP-KG-Retrain} retrains the full
model after each batch as an upper-bound reference. BMGC-Retrain and
DEMM-Retrain are additionally included to contextualize the retraining cost of
competitive transductive baselines. For ACM, DBLP, and YELP, 200 nodes are
held out in 2 batches of 100; for MAG, 2\% of nodes (2,278 samples)
are withheld in 5 batches, simulating a high-throughput streaming scenario.

Table~\ref{tab:inductive_exp} reports per-batch ACC, F1, and inference time.
The efficiency advantage is decisive: on MAG, BMGC-Retrain and DEMM-Retrain
require \textbf{over 65 seconds per batch}, whereas SNAP-KG-Projector completes the same assignment in
\textbf{milliseconds}, an multiple orders-of-magnitude speedup that makes continuous
entity integration practical at scale. Crucially, this comes at no quality
cost: SNAP-KG-Projector consistently outperforms both Isolated and Neighbors,
and its ACC and F1 on unseen nodes closely match the retrain upper bound in
Table~\ref{tab:table_benchmark} across ACM, DBLP, and YELP; on MAG the
projector slightly exceeds it, likely due to larger cluster granularity reducing
assignment ambiguity at scale.

\begin{figure}[H]
    \centering
    \includegraphics[width=0.62\textwidth]{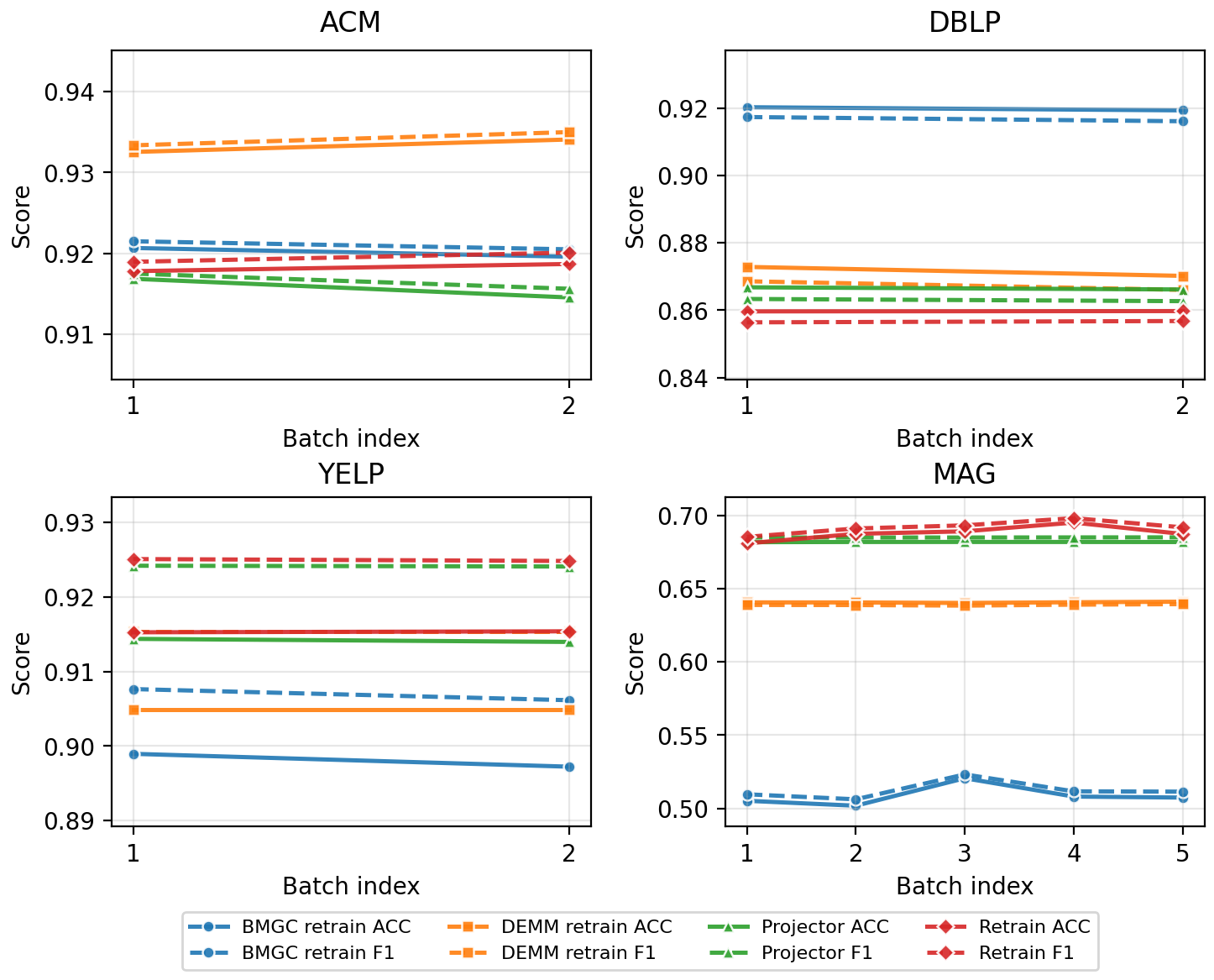}
    \caption{Overall ACC and F1 across successive batches on ACM, DBLP, YELP,
        and MAG. Solid: ACC; dashed: F1. SNAP-KG-Projector (green) closely
        tracks SNAP-KG-Retrain (red), while BMGC-Retrain (blue) and
        DEMM-Retrain (orange) require $>$65\,s of retraining per batch on MAG.}
    \label{fig:inductive_acc_f1_four}
\end{figure}

Figure~\ref{fig:inductive_acc_f1_four} shows overall ACC and F1 after
integrating each batch across all four datasets. SNAP-KG-Projector (green)
closely tracks SNAP-KG-Retrain (red) throughout, while BMGC-Retrain (blue) and
DEMM-Retrain (orange) incur over 65 seconds of retraining per batch on MAG to
achieve comparable quality. The projector thus provides retrain-level quality
at a fraction of the cost.

\begin{table}[!htb]
    \centering
    \small
    \caption{Per-batch inductive evaluation. ACC (\%), F1 (\%), and inference
        time (s) per batch. SNAP-KG-Retrain, BMGC-Retrain, and DEMM-Retrain
        report runtime only. \textbf{SNAP-KG-Projector achieves
        multiple orders-of-magnitude speedups} (milliseconds vs.\ $>$65\,s per batch on
        MAG) while matching retrain-level quality.}
    \label{tab:inductive_exp}
    \begin{adjustbox}{max width=\textwidth}
\begin{tabular}{l|l|rrr|rrr|rrr|r|r|r}
\hline
 &
   &
  \multicolumn{3}{l|}{Isolated} &
  \multicolumn{3}{l|}{Neighbors} &
  \multicolumn{3}{l|}{SNAP-KG-Projector} &
  \multicolumn{1}{l|}{\begin{tabular}[c]{@{}l@{}}SNAP-KG-\\ Retrain\end{tabular}} &
  \multicolumn{1}{l|}{\begin{tabular}[c]{@{}l@{}}BMGC-\\ Retrain\end{tabular}} &
  \multicolumn{1}{l}{\begin{tabular}[c]{@{}l@{}}DEMM-\\ Retrain\end{tabular}} \\ \cline{3-14} 
 &
  \begin{tabular}[c]{@{}l@{}}Batch / \\ Nodes\end{tabular} &
  \multicolumn{1}{l}{ACC (\%)} &
  \multicolumn{1}{l}{F1 (\%)} &
  \multicolumn{1}{l|}{Time (s)} &
  \multicolumn{1}{l}{ACC (\%)} &
  \multicolumn{1}{l}{F1 (\%)} &
  \multicolumn{1}{l|}{Time (s)} &
  \multicolumn{1}{l}{ACC (\%)} &
  \multicolumn{1}{l}{F1 (\%)} &
  \multicolumn{1}{l|}{Time (s)} &
  \multicolumn{1}{l|}{Time (s)} &
  \multicolumn{1}{l|}{Time (s)} &
  \multicolumn{1}{l}{Time (s)} \\ \hline
\multirow{2}{*}{ACM}  & 1 / 100 & 93.66 & 81.83 & 0.008 & 96.00 & 88.24 & 0.071 & 97.33 & 93.13 & \textbf{0.003} & 187.176    & 167.509 & 5.853   \\
                      & 2 / 100 & 88.49 & 68.67 & 0.013 & 90.83 & 74.13 & 0.102 & 91.00 & 74.97 & \textbf{0.004} & 187.711    & 166.111 & 6.473   \\ \hline
\multirow{2}{*}{DBLP} & 1 / 100 & 71.66 & 70.31 & 0.009 & 77.33 & 76.31 & 0.083 & 80.00 & 78.72 & \textbf{0.003} & 290.011    & 220.960 & 17.711  \\
                      & 2 / 100 & 73.50 & 72.52 & 0.007 & 79.33 & 78.72 & 0.087 & 82.00 & 81.62 & \textbf{0.002} & 313.057    & 189.786 & 19.119  \\ \hline
\multirow{2}{*}{YELP} & 1 / 100 & 85.33 & 85.09 & 0.007 & 90.33 & 89.75 & 0.038 & 92.00 & 91.47 & \textbf{0.005} & 140.273    & 85.635  & 7.905   \\
                      & 2 / 100 & 82.33 & 82.22 & 0.015 & 86.49 & 86.56 & 0.046 & 91.16 & 92.47 & \textbf{0.005} & 120.522    & 88.111  & 8.228   \\ \hline
\multirow{5}{*}{MAG}  & 1 / 500 & 61.60 & 59.97 & 0.023 & 67.40 & 65.89 & 4.899 & 71.20 & 69.93 & \textbf{0.021} & 10,175.000 & 226.327 & 114.814 \\
                      & 2 / 500 & 63.90 & 60.99 & 0.018 & 66.90 & 64.63 & 0.476 & 71.20 & 68.90 & \textbf{0.003} & 8,565.156  & 139.139 & 84.705  \\
                      & 3 / 500 & 63.06 & 61.12 & 0.015 & 66.13 & 64.63 & 0.462 & 70.23 & 68.63 & \textbf{0.004} & 9,928.550  & 134.030 & 83.307  \\
                      & 4 / 500 & 62.55 & 61.49 & 0.013 & 66.50 & 65.74 & 0.465 & 69.65 & 68.79 & \textbf{0.003} & 11,617.698 & 221.615 & 126.326 \\
                      & 5 / 278 & 62.11 & 61.51 & 0.009 & 65.94 & 65.54 & 0.249 & 69.23 & 68.87 & \textbf{0.006} & 10,171.021 & 132.789 & 84.406  \\ \hline
\end{tabular}%

    \end{adjustbox}
\end{table}

\subsection{Clustering Quality: Inductive vs.\ Transductive}
\label{sec:MVGC}
Table~\ref{tab:table_benchmark} contextualizes SNAP-KG's clustering quality
against transductive state-of-the-art baselines. For fair comparison, all
methods are trained on the complete node set of each dataset; all baselines
additionally require this complete graph at inference time and cannot embed
unseen nodes without full retraining, whereas SNAP-KG's inductive inference
is evaluated separately in Section~\ref{sec:inductive_exp}.

SNAP-KG achieves competitive performance across all datasets. On MAG, the
largest benchmark with 113,919 nodes, it attains the \textbf{best ACC
(68.20\%)} and \textbf{F1 (68.35\%)} among all methods. On ACM, DBLP, and YELP, ACC
reaches 91.50\%, 91.82\%, and 91.20\% respectively, within 2\% of the
strongest transductive baselines. On IMDB, all methods achieve low NMI and ARI scores regardless of approach,
suggesting that the graph structure itself provides limited discriminative
signal for clustering on this dataset.
\begin{table}[!htb]
    \centering
    \small
    \caption{Clustering on standard multi-view graph benchmarks. All methods use the
        complete node set. Best results are \textbf{bolded}; second-best are
        \underline{underlined}. All baselines require full retraining for
        unseen nodes; SNAP-KG does not.}
    \label{tab:table_benchmark}
    \begin{adjustbox}{max width=\textwidth}
\begin{tabular}{lrrrrrrrrllll}
\hline
\multicolumn{1}{c}{\multirow{2}{*}{Method}} &
  \multicolumn{4}{c}{ACM} &
  \multicolumn{4}{c}{DBLP} &
  \multicolumn{4}{c}{IMDB} \\ \cline{2-13} 
\multicolumn{1}{c}{} &
  \multicolumn{1}{l}{NMI (\%)} &
  \multicolumn{1}{l}{ARI (\%)} &
  \multicolumn{1}{l}{ACC (\%)} &
  \multicolumn{1}{l}{F1 (\%)} &
  \multicolumn{1}{l}{NMI (\%)} &
  \multicolumn{1}{l}{ARI (\%)} &
  \multicolumn{1}{l}{ACC (\%)} &
  \multicolumn{1}{l}{F1 (\%)} &
  NMI (\%) &
  ARI (\%) &
  ACC (\%) &
  F1 (\%) \\ \hline
MGDCR~\cite{mo2023multiplex} &
  65.88 &
  70.39 &
  88.96 &
  89.08 &
  71.47 &
  76.71 &
  90.19 &
  90.19 &
  \multicolumn{1}{r}{\textbf{8.93}} &
  \multicolumn{1}{r}{9.01} &
  \multicolumn{1}{r}{52.34} &
  \multicolumn{1}{r}{\textbf{49.34}} \\
AGE~\cite{cui2020adaptive} &
  65.51 &
  70.07 &
  88.98 &
  88.95 &
  56.92 &
  58.37 &
  81.88 &
  81.63 &
  \multicolumn{1}{r}{7.26} &
  \multicolumn{1}{r}{\textbf{12.13}} &
  \multicolumn{1}{r}{\textbf{55.15}} &
  \multicolumn{1}{r}{{\ul{45.67}}} \\
BMGC~\cite{shen2024balanced} &
  \textbf{76.53} &
  \textbf{81.03} &
  \textbf{93.26} &
  \textbf{93.28} &
  77.13 &
  82.41 &
  92.75 &
  92.36 &
  \multicolumn{1}{r}{{\ul{7.28}}} &
  \multicolumn{1}{r}{6.49} &
  \multicolumn{1}{r}{45.19} &
  \multicolumn{1}{r}{41.52} \\
DMG~\cite{mo2023disentangled} &
  75.38 &
  80.00 &
  92.91 &
  92.97 &
  77.45 &
  82.73 &
  92.93 &
  92.55 &
  \multicolumn{1}{r}{6.35} &
  \multicolumn{1}{r}{9.72} &
  \multicolumn{1}{r}{47.50} &
  \multicolumn{1}{r}{33.91} \\
DuaLGR~\cite{ling2023dual} &
  69.18 &
  74.67 &
  90.74 &
  90.73 &
  {\ul{78.06}} &
  {\ul{83.76}} &
  {\ul{93.25}} &
  {\ul{92.73}} &
  \multicolumn{1}{r}{4.97} &
  \multicolumn{1}{r}{{\ul{9.99}}} &
  \multicolumn{1}{r}{52.72} &
  \multicolumn{1}{r}{42.45} \\
O2MAC~\cite{fan2020one2multi} &
  61.82 &
  65.55 &
  86.74 &
  86.92 &
  70.38 &
  75.10 &
  89.62 &
  89.01 &
  \multicolumn{1}{r}{3.02} &
  \multicolumn{1}{r}{3.96} &
  \multicolumn{1}{r}{47.68} &
  \multicolumn{1}{r}{41.25} \\
DEMM~\cite{lin2025effective} &
  {\ul{75.45}} &
  {\ul{80.43}} &
  {\ul{93.09}} &
  {\ul{93.17}} &
  \textbf{79.07} &
  \textbf{84.55} &
  \textbf{93.67} &
  \textbf{93.29} &
  \multicolumn{1}{r}{2.53} &
  \multicolumn{1}{r}{4.45} &
  \multicolumn{1}{r}{{\ul{54.94}}} &
  \multicolumn{1}{r}{31.77} \\
SNAP-KG (ours) &
  71.08 &
  76.25 &
  91.50 &
  91.59 &
  74.03 &
  80.31 &
  91.82 &
  91.34 &
  \multicolumn{1}{r}{3.42} &
  \multicolumn{1}{r}{3.48} &
  \multicolumn{1}{r}{46.67} &
  \multicolumn{1}{r}{36.01} \\ \hline
\multicolumn{1}{c}{\multirow{2}{*}{Method}} &
  \multicolumn{4}{c}{YELP} &
  \multicolumn{4}{c}{MAG} &
   &
   &
   &
   \\ \cline{2-9}
\multicolumn{1}{c}{} &
  \multicolumn{1}{l}{NMI (\%)} &
  \multicolumn{1}{l}{ARI (\%)} &
  \multicolumn{1}{l}{ACC (\%)} &
  \multicolumn{1}{l}{F1 (\%)} &
  \multicolumn{1}{l}{NMI (\%)} &
  \multicolumn{1}{l}{ARI (\%)} &
  \multicolumn{1}{l}{ACC (\%)} &
  \multicolumn{1}{l}{F1 (\%)} &
   &
   &
   &
   \\ \cline{1-9}
MGDCR~\cite{mo2023multiplex} &
  38.44 &
  42.11 &
  64.96 &
  64.96 &
  {\ul{57.20}} &
  45.06 &
  58.69 &
  58.96 &
   &
   &
   &
   \\
AGE~\cite{cui2020adaptive} &
  64.52 &
  66.77 &
  88.91 &
  90.91 &
  59.79 &
  \textbf{49.68} &
  {\ul{66.91}} &
  {\ul{64.85}} &
   &
   &
   &
   \\
BMGC~\cite{shen2024balanced} &
  {\ul{70.18}} &
  72.41 &
  90.95 &
  91.86 &
  \textbf{57.62} &
  {\ul{48.18}} &
  64.38 &
  62.45 &
   &
   &
   &
   \\
DMG~\cite{mo2023disentangled} &
  38.81 &
  41.78 &
  63.80 &
  60.62 &
  36.43 &
  29.23 &
  53.98 &
  53.54 &
   &
   &
   &
   \\
DuaLGR~\cite{ling2023dual} &
  66.12 &
  70.61 &
  90.17 &
  90.87 &
  44.80 &
  38.19 &
  62.40 &
  62.50 &
   &
   &
   &
   \\
O2MAC~\cite{fan2020one2multi} &
  39.02 &
  42.53 &
  65.07 &
  56.74 &
  36.49 &
  31.25 &
  53.37 &
  45.53 &
   &
   &
   &
   \\
DEMM~\cite{lin2025effective} &
  \textbf{72.44} &
  \textbf{76.54} &
  \textbf{92.40} &
  \textbf{93.01} &
  56.87 &
  46.13 &
  64.49 &
  63.99 &
   &
   &
   &
   \\
SNAP-KG (ours) &
  69.65 &
  {\ul{72.95}} &
  {\ul{91.20}} &
  {\ul{92.16}} &
  50.85 &
  43.45 &
  \textbf{68.20} &
  \textbf{68.35} &
   &
   &
   &
   \\[2pt] \cline{1-9}
\end{tabular}%
    \end{adjustbox}
\end{table}

\subsection{Ablation Study}
\label{sec:ablation}
To validate the contribution of multi-view fusion, we compare the full SNAP-KG
model against single-view variants trained on each adjacency matrix in
isolation, with SPARC and SimMLP as per-view spectral and MLP baselines. The result is shown in Table~\ref{tab:one_view}. The full model consistently outperforms every
single-view variant: on ACM, fusing PAP and PLP raises NMI from 41.28\% to
71.08\%; on DBLP, the weakest single view (APA) achieves only 5.02\% NMI
against 74.03\% for the full model. Consistent gains on YELP and MAG confirm
that multi-view aggregation is necessary rather than redundant.


\begin{table}[!htb]
    \centering
    \small
    \caption{Single-view ablation. Each row reports clustering performance when
             SNAP-KG uses only the indicated view; ``All views'' is the full
             multi-view model.}
    \label{tab:one_view}
    \begin{adjustbox}{max width=\textwidth}
\begin{tabular}{lllrrrrlllrrrr}
\hline
\multicolumn{3}{l}{} &
  \multicolumn{1}{l}{NMI (\%)} &
  \multicolumn{1}{l}{ARI (\%)} &
  \multicolumn{1}{l}{ACC (\%)} &
  \multicolumn{1}{l}{F1 (\%)} &
   &
   &
   &
  \multicolumn{1}{l}{NMI (\%)} &
  \multicolumn{1}{l}{ARI (\%)} &
  \multicolumn{1}{l}{ACC (\%)} &
  \multicolumn{1}{l}{F1 (\%)} \\ \hline
\multirow{7}{*}{ACM} &
  \multirow{3}{*}{SNAP-KG} &
  All views &
  71.08 &
  76.25 &
  91.5 &
  91.59 &
  \multirow{7}{*}{IMDB} &
  \multirow{3}{*}{SNAP-KG} &
  All views &
  3.42 &
  3.48 &
  46.67 &
  36.01 \\
 &
   &
  PAP &
  27.86 &
  22.18 &
  54.94 &
  56.48 &
   &
   &
  MAM &
  0.98 &
  0.04 &
  40.5 &
  36.01 \\
 &
   &
  PLP &
  41.28 &
  34.53 &
  65.02 &
  66.33 &
   &
   &
  MDM &
  0.34 &
  0.008 &
  35.71 &
  34.59 \\ \cline{2-7} \cline{9-14} 
 &
  \multirow{2}{*}{SPARC} &
  PAP &
  44.52 &
  51.61 &
  81.16 &
  88.15 &
   &
  \multirow{2}{*}{SPARC} &
  MAM &
  1.75 &
  4.56 &
  45.92 &
  39.98 \\
 &
   &
  PLP &
  58.32 &
  64.84 &
  86.88 &
  86.79 &
   &
   &
  MDM &
  0.45 &
  1.54 &
  43.87 &
  36.82 \\ \cline{2-7} \cline{9-14} 
 &
  \multirow{2}{*}{SimMLP} &
  PAP &
  59.34 &
  65.09 &
  87.14 &
  87.14 &
   &
  \multirow{2}{*}{SimMLP} &
  MAM &
  1.23 &
  2.07 &
  44.52 &
  36.81 \\
 &
   &
  PLP &
  45.64 &
  49.1 &
  78.98 &
  78.95 &
   &
   &
  MDM &
  1.53 &
  4.08 &
  47.36 &
  38.25 \\ \hline
\multirow{10}{*}{DBLP} &
  \multirow{4}{*}{SNAP-KG} &
  All views &
  74.03 &
  80.31 &
  91.82 &
  91.34 &
  \multirow{10}{*}{YELP} &
  \multirow{4}{*}{SNAP-KG} &
  All views &
  69.65 &
  72.95 &
  91.2 &
  92.16 \\
 &
   &
  APTPA &
  21.51 &
  20.66 &
  56.42 &
  55.9 &
   &
   &
  BLB &
  40.27 &
  31.28 &
  70.08 &
  74.51 \\
 &
   &
  APVPA &
  71.12 &
  76.78 &
  90.07 &
  89.4 &
   &
   &
  BSB &
  66.58 &
  69.82 &
  90.05 &
  91.03 \\
 &
   &
  APA &
  5.02 &
  4.58 &
  34.98 &
  34.46 &
   &
   &
  BUB &
  0.57 &
  0.71 &
  37.18 &
  35.82 \\ \cline{2-7} \cline{9-14} 
 &
  \multirow{3}{*}{SPARC} &
  APTPA &
  40.68 &
  30.35 &
  60.74 &
  61.9 &
   &
  \multirow{3}{*}{SPARC} &
  BLB &
  20.85 &
  20.93 &
  63.35 &
  63.13 \\
 &
   &
  APVPA &
  66.12 &
  65.78 &
  84.82 &
  85.92 &
   &
   &
  BSB &
  34.45 &
  28.03 &
  69.89 &
  73.01 \\
 &
   &
  APA &
  3.43 &
  2.5 &
  32.88 &
  31.69 &
   &
   &
  BUB &
  36.97 &
  29.37 &
  69.32 &
  73.24 \\ \cline{2-7} \cline{9-14} 
 &
  \multirow{3}{*}{SimMLP} &
  APTPA &
  10.69 &
  10.23 &
  31.08 &
  28.45 &
   &
  \multirow{3}{*}{SimMLP} &
  BLB &
  42.93 &
  36.62 &
  74.56 &
  77.91 \\
 &
   &
  APVPA &
  43.58 &
  40.72 &
  65.66 &
  64.85 &
   &
   &
  BSB &
  43.43 &
  39.97 &
  76.89 &
  79.33 \\
 &
   &
  APA &
  30.81 &
  26.77 &
  53.44 &
  54.62 &
   &
   &
  BUB &
  37.21 &
  28.66 &
  70.39 &
  73.21 \\ \hline
\multirow{7}{*}{MAG} &
  \multirow{3}{*}{SNAP-KG} &
  All views &
  50.85 &
  43.45 &
  68.2 &
  68.35 &
   &
   &
   &
  \multicolumn{1}{l}{} &
  \multicolumn{1}{l}{} &
  \multicolumn{1}{l}{} &
  \multicolumn{1}{l}{} \\
 &
   &
  PP &
  37.2 &
  32.98 &
  62 &
  61.97 &
   &
   &
   &
  \multicolumn{1}{l}{} &
  \multicolumn{1}{l}{} &
  \multicolumn{1}{l}{} &
  \multicolumn{1}{l}{} \\
 &
   &
  PAP &
  37.72 &
  34.06 &
  62.12 &
  62.36 &
   &
   &
   &
  \multicolumn{1}{l}{} &
  \multicolumn{1}{l}{} &
  \multicolumn{1}{l}{} &
  \multicolumn{1}{l}{} \\ \cline{2-7}
 &
  \multirow{2}{*}{SPARC} &
  PP &
  32.17 &
  26.18 &
  54.77 &
  52.21 &
   &
   &
   &
  \multicolumn{1}{l}{} &
  \multicolumn{1}{l}{} &
  \multicolumn{1}{l}{} &
  \multicolumn{1}{l}{} \\
 &
   &
  PAP &
  45.97 &
  36.58 &
  54.84 &
  50.7 &
   &
   &
   &
  \multicolumn{1}{l}{} &
  \multicolumn{1}{l}{} &
  \multicolumn{1}{l}{} &
  \multicolumn{1}{l}{} \\ \cline{2-7}
 &
  \multirow{2}{*}{SimMLP} &
  PP &
  49.03 &
  39.53 &
  61.39 &
  61.51 &
   &
   &
   &
  \multicolumn{1}{l}{} &
  \multicolumn{1}{l}{} &
  \multicolumn{1}{l}{} &
  \multicolumn{1}{l}{} \\
 &
   &
  PAP &
  49.81 &
  40.04 &
  62.21 &
  62.57 &
   &
   &
   &
  \multicolumn{1}{l}{} &
  \multicolumn{1}{l}{} &
  \multicolumn{1}{l}{} &
  \multicolumn{1}{l}{} \\ \cline{1-7}
\end{tabular}%
    \end{adjustbox}
\end{table}

\subsection{Entity Resolution \& Link Prediction}
\label{sec:ER_LP}


Both experiments follow the same inductive partition as
Section~\ref{sec:inductive_exp}. For ER, since the multi-view graph benchmarks
do not provide ground-truth duplicate pairs, we construct synthetic duplicates
by injecting feature-proportional Gaussian noise ($\boldsymbol{\epsilon} \sim
\mathcal{N}(0,\,\sigma^{2}\,\mathrm{diag}(\hat{\sigma}_{x}^{2}))$,
$\sigma{=}0.15$) into existing node features (50\% of each batch), with match
decisions at cosine similarity threshold 0.90; DBLP-ACM is additionally
evaluated as a real-world benchmark with genuine duplicate pairs. For LP, the
task is binary edge existence prediction given a new and a candidate node
without relation type; ground-truth cross-edges are derived from the full graph.
SNAP-KG introduces two lightweight task-specific heads initialized from the
frozen $f_{\phi}$ and fine-tuned on existing nodes only: an \emph{entity
resolver} and an \emph{LP head}; because $f_{\phi}$ remains frozen, cluster
assignments are unaffected by task-specific adaptation. We compare
\textbf{SNAP-KG (clustering)}, restricting candidates to the assigned cluster,
against \textbf{Full} (entire node set); \textbf{Reduction (\%)} measures the
decrease in average candidates per node.

SNAP-KG achieves \textbf{62--75\% candidate reduction} while preserving ER
quality close to the full-scan upper bound, as shown in Table~\ref{tab:ER_LP_with_DBLP_ACM}:
on ACM, precision, recall, and F1 remain perfect at 100\% despite 65.4\%
reduction. Intra-cluster restriction can further improve LP by filtering false
positives; on DBLP, SNAP-KG outperforms full-scan in AUPR (83.01\% vs.\
75.75\%), but true positive edges spanning cluster boundaries are excluded,
contributing to moderate LP degradation on MAG, compounded by roughly 60\% inductive
clustering accuracy, as shown in Table~\ref{tab:inductive_exp}. This asymmetry
reflects each task's nature: ER duplicates share nearly identical features and
co-locate within the same cluster; LP edges connect structurally related but
semantically distinct nodes whose embeddings may span different clusters, making
LP more susceptible to boundary exclusions. To validate beyond synthetic noise, we additionally evaluate on
DBLP-ACM~\cite{kopcke2010evaluation}, a real-world cross-dataset
benchmark in which the model is trained on DBLP and streaming
entities arrive from ACM, a strictly inductive transfer setting
with no overlap between training and inference graphs. With $K{=}5$,
selected as the best-performing configuration across a range of
$K$ values, SNAP-KG achieves 76.2\% candidate reduction while
preserving F1 within 0.81 percentage points of full-scan (92.64\% vs.\ 93.45\%),
confirming generalization to real-world heterogeneous ER under
genuine domain shift.


\paragraph{LP at cluster boundaries.}
On MAG, scoping drops 14{,}938 of the 75{,}578 ground-truth positive edges, so
\textbf{19.8\% of true edges cross a cluster boundary}, and 45 of 2{,}254
arriving nodes (2.0\%) have no positive edge left. That is the real cost of
scoping. But it is not what Table~\ref{tab:ER_LP_with_DBLP_ACM} shows, since an
edge outside the candidate set is never scored. The drop in the table comes from
the negatives instead. They now all come from inside the cluster, so they are
much harder to separate from real edges than the random nodes a full scan picks,
and AUC-ROC falls from 96.43\% to 92.21\%. AUPR has one more issue. It depends
on what fraction of the scored pairs are true edges, and that fraction shrinks
because there are fewer true edges but about the same number of scored pairs.
Part of the drop from 61.00\% to 37.95\% comes from this, not from worse
ranking.


\begin{table}[!htb]
    \centering
    \small
    \caption{ER and LP under intra-cluster scoping (SNAP-KG) vs.\ full-scan
        (Full). Reduction (\%) measures average candidate decrease per node.
        DBLP-ACM is a real-world cross-dataset ER benchmark; LP not applicable.}
    \label{tab:ER_LP_with_DBLP_ACM}
    \begin{adjustbox}{max width=\textwidth}
\begin{tabular}{crrrrrrrr}
\hline
\multirow{2}{*}{Dataset} &
  \multicolumn{1}{c}{\multirow{2}{*}{Type}} &
  \multicolumn{1}{c}{\multirow{2}{*}{\begin{tabular}[c]{@{}c@{}}Average \\ Candidate\end{tabular}}} &
  \multicolumn{1}{c}{\multirow{2}{*}{\begin{tabular}[c]{@{}c@{}}Search \\ Reduction (\%)\end{tabular}}} &
  \multicolumn{3}{c}{Entity Resolution} &
  \multicolumn{2}{c}{Link Prediction} \\ \cline{5-9} 
 &
  \multicolumn{1}{c}{} &
  \multicolumn{1}{c}{} &
  \multicolumn{1}{c}{} &
  \multicolumn{1}{c}{Precision (\%)} &
  \multicolumn{1}{c}{Recall (\%)} &
  \multicolumn{1}{c}{F1 (\%)} &
  \multicolumn{1}{c}{AUC-ROC (\%)} &
  \multicolumn{1}{c}{AUPR (\%)} \\ \hline
\multirow{2}{*}{ACM}  & SNAP-KG (clustering) & 1026.5  & 65.4 & 100   & 100   & 100   & 73.30  & 41.43 \\
                      & Full                 & 2964    & -    & 100   & 100   & 100   & 81.11 & 50.71 \\
\multirow{2}{*}{DBLP} & SNAP-KG (clustering) & 1013.7  & 74.5 & 71.80 & 88.80 & 79.30 & 71.18  & 83.01 \\
                      & Full                 & 3977    & -    & 71.80 & 91.20 & 80.30 & 76.20  & 75.75 \\
\multirow{2}{*}{IMDB} & SNAP-KG (clustering) & 1625.3  & 65.3 & 92.10 & 95    & 93.50 & 59.57  & 26.75 \\
                      & Full                 & 4684    & -    & 92.50 & 100   & 96    & 64.67  & 32.86 \\
\multirow{2}{*}{YELP} & SNAP-KG (clustering) & 953.2   & 62.7 & 58.20 & 70    & 63.70 & 81.63  & 62.42 \\
                      & Full                 & 2554    & -    & 58.30 & 71.70 & 64.30 & 87.04  & 68.64 \\
\multirow{2}{*}{MAG}  & SNAP-KG (clustering) & 28164.5 & 74.8 & 81.40 & 97.90 & 88.80 & 92.21  & 37.95 \\
                      & Full                 & 111640  & -    & 81.40 & 99.60 & 89.60 & 96.43  & 61    \\ \hline
\multirow{2}{*}{\begin{tabular}[c]{@{}c@{}}DBLP-ACM\\ (Cross-dataset)\end{tabular}} &
  \begin{tabular}[c]{@{}r@{}}SNAP-KG (clustering)\\ (K=5)\end{tabular} &
  623.8 &
  76.15 &
  93.02 &
  92.26 &
  92.64 &
  - &
  - \\
                      & Full                 & 2614    & -    & 93.24 & 93.66 & 93.45 & -      & -     \\ \hline
\end{tabular}
    \end{adjustbox}
\end{table}


\subsection{Production-Scale Knowledge Graph: OGB-WikiKG2}
\label{sec:wikikg2}
To assess scalability, we evaluate SNAP-KG on OGB-WikiKG2~\cite{hu2020open}
with 2.5M entities and 535 relation types. Due to computational constraints,
views are constructed from the 8 highest-frequency relations (\textit{instance
of}, \textit{occupation}, \textit{sex or gender}, \textit{given name},
\textit{country of citizenship}, \textit{place of birth}, \textit{cast member},
\textit{country}), retaining 2.4M nodes after preprocessing. For ER, 800
synthetic query nodes are constructed by injecting Gaussian noise following the
protocol in Section~\ref{sec:ER_LP}; for LP, 1,000 held-out nodes are used. The
full-scan baseline is approximated by sampling 500,000 nodes uniformly at random
to avoid out-of-memory conditions; the cluster-scoped variant compares each new
node against all nodes within its assigned cluster under the same cap. In
practice, no cluster exceeded this cap, confirming well-balanced partitions at
this scale. Because the full-scan baseline already operates on a sampled subset,
it constitutes a relaxed upper bound on recall, making any observed performance
gap a conservative estimate of SNAP-KG's advantage.

Since OGB-WikiKG2 carries no ground-truth community annotations, $K$ is treated
as a hyperparameter over $\{200, 225, 250, 275\}$, selected by downstream
performance on a held-out validation subset. $K{=}250$ yields the best results
for both tasks; notably, LP AUPR under cluster-scoped search exceeds the
full-scan baseline by by 17.78, as shown in Table~\ref{tab:wikikg2}, consistent with
the noise-filtering effect observed on smaller benchmarks.

\paragraph{Scaling to more predicates.}
Training cost grows with the number of views $M$. Each view has its own GNN, so
that part is linear in $M$, and the fusion encoder adds $\mathcal{O}(M^{2})$
attention over $M{+}1$ tokens. Inference cost does not grow at all, because the
projector only reads $\boldsymbol{x}_{\mathrm{new}}$ and never touches a view.
Per-entity latency is the same whatever $M$ is, so the eight-predicate limit
above is a training-budget choice, not an architectural one.

\begin{table}[!htb]
    \centering
    \small
    \caption{ER and LP on OGB-WikiKG2. Full-scan baseline uses a 500,000-node
        random sample; cluster-scoped results restrict search to the assigned
        cluster.}
    \label{tab:wikikg2}
    \begin{adjustbox}{max width=\textwidth}
\begin{tabular}{rrrrrrrr}
\hline
\multicolumn{3}{c}{}                        & \multicolumn{3}{c}{Entity Resolution} & \multicolumn{2}{c}{Link Prediction} \\ \hline
Type &
  \multicolumn{1}{c}{\begin{tabular}[c]{@{}c@{}}Average \\ Candidate\end{tabular}} &
  \multicolumn{1}{c}{\begin{tabular}[c]{@{}c@{}}Search \\ Reduction (\%)\end{tabular}} &
  \multicolumn{1}{c}{Precision (\%)} &
  \multicolumn{1}{c}{Recall (\%)} &
  \multicolumn{1}{c}{F1 (\%)} &
  \multicolumn{1}{c}{AUC-ROC (\%)} &
  \multicolumn{1}{c}{AUPR (\%)} \\ \hline
SNAP-KG (clustering), K=200 & 13478  & 97.30 & 86.72       & 73.52      & 79.58      & 98.96            & 19.19            \\
SNAP-KG (clustering), K=225 & 12833  & 97.43 & 88.98       & 71.62      & 79.36      & 98.35            & 15.84            \\
SNAP-KG (clustering), K=250 & 11230  & 97.75 & 90.51       & 72.75      & 80.67      & 98.24            & 31.13            \\
SNAP-KG (clustering), K=275 & 10435  & 97.91 & 89.66       & 72.62      & 80.25      & 98.10            & 30.35            \\
Full                       & 500000 & -     & 80.50       & 80.50      & 80.50      & 97.75            & 13.35            \\ \hline
\end{tabular}%

    \end{adjustbox}
\end{table}

\section{Conclusion}
We presented SNAP-KG, the first \emph{graph-structural} multi-view framework
to simultaneously support relational clustering and inductive inference for
streaming KG construction.
By distilling a fused GNN-Transformer pipeline into a lightweight MLP
projector, SNAP-KG reduces per-entity inference from a full retraining
operation to a single feed-forward pass: the inductive projector matches
retrain-level cluster assignment quality using only raw node features at
inference time, with no graph access and no retraining, achieving
multiple orders-of-magnitude speedups over retraining-based baselines.
Experiments across five benchmark datasets and a 2.4M-node production KG confirm
62--75\% candidate search reduction on the benchmark datasets and
97\% on OGB-WikiKG2 candidate search reduction for entity resolution and
link prediction scoping, with ER quality preserved close
to the full-scan upper bound and validated inductively on a real-world
cross-dataset benchmark (DBLP-ACM), where the model is trained on DBLP
and generalises to unseen ACM entities under genuine domain shift.
The framework is agnostic to the choice of clustering algorithm; because
re-clustering is far cheaper than retraining, practitioners can optionally
refresh cluster structure to adapt to evolving graph distributions without
modifying model weights. The graph-aware cluster centroids are further
composable with IVF-style ANN indices~\cite{johnson2019billion} for
sub-linear retrieval at production scale; we leave the empirical evaluation
of such composed pipelines to future work.


%
%

\newpage
\subsubsection*{Declaration of use of Generative AI}
We used a large language model (Claude, Anthropic) solely to assist
with grammar correction and writing refinement. All ideas, problem formulation,
methodology, experimental design, results, and conclusions are entirely 
our own. We take full responsibility for the accuracy and
integrity of all content presented in this work.

\subsubsection*{Supplemental Material Statement}
The source code, datasets, and detailed experimental configurations used in this work are available at: \url{https://github.com/ryanjclin/SNAP-KG}.

\bibliographystyle{splncs04}
\bibliography{references}

\end{document}